\documentclass[review,12pt,authoryear]{elsarticle}
\journal{}
\biboptions{sort&compress}
\graphicspath{{Images/}}
\usepackage{xcolor}

\usepackage{setspace}
\usepackage{tablefootnote}
\usepackage{hyperref}

\usepackage{amsmath}
\usepackage{amsfonts}
\usepackage{amssymb}
\usepackage{booktabs}
\usepackage{siunitx}
\usepackage{times}
\usepackage{cleveref}
\usepackage{array}
\newcolumntype{P}[1]{>{\centering\arraybackslash}p{#1}}

\begin{document}

\begin{frontmatter}

\title{Muzzle-Based Cattle Identification System Using Artificial Intelligence (AI)}

\author[]{Hasan~Zohirul~Islam\corref{cor1}\fnref{fn}}
% \address[label1]{adorsho praniSheba Ltd, Mohakhali C/A, Dhaka-1212, Bangladesh}
\ead{hzihimel@gmail.com}
\cortext[cor1]{Corresponding author.}
\fntext[fn]{Work performed while at adorsho praniSheba Ltd.}

\RenewDocumentCommand{\footnote}{ m }{\ogfootnote{\onehalfspacing{}#1}}

\author[]{Safayet~Khan\fnref{fn}}
\ead{ratulme12@gmail.com}

\author[label1]{Sanjib~Kumar~Paul}
\ead{sanjibpaul171@gmail.com}

\author[]{Sheikh~Imtiaz~Rahi\fnref{fn}}
\ead{imtiaz.rahi@gmail.com}

\author[label1]{Fahim~Hossain~Sifat}
\ead{fahimsifat29@gmail.com}

\author[]{Md.~Mahadi~Hasan~Sany\fnref{fn}}
\ead{mahadi15-11173@diu.edu.bd}

\author[]{Md.~Shahjahan~Ali~Sarker\fnref{fn}}
\ead{shahjahansarker1981@gmail.com} 

\author[]{Tareq~Anam\fnref{fn}}
\ead{tareqanam@outlook.com}

\author[]{Ismail~Hossain~Polas\fnref{fn}}
\ead{ismailhossainpolash@gmail.com}

\begin{abstract}
Absence of tamper-proof cattle identification technology was a significant problem preventing insurance companies from providing livestock insurance. This lack of technology had devastating financial consequences for marginal farmers as they did not have the opportunity to claim compensation for any unexpected events such as the accidental death of cattle in Bangladesh. Using machine learning and deep learning algorithms, we have solved the bottleneck of cattle identification by developing and introducing a muzzle-based cattle identification system. The uniqueness of cattle muzzles has been scientifically established, which resembles human fingerprints. This is the fundamental premise that prompted us to develop a cattle identification system that extracts the uniqueness of cattle muzzles. For this purpose, we collected 32,374 images from 826 cattle. Contrast-limited adaptive histogram equalization (CLAHE) with sharpening filters was applied in the preprocessing steps to remove noise from images. We used the YOLO algorithm for cattle muzzle detection in the image and the FaceNet architecture to learn unified embeddings from muzzle images using squared $L_2$ distances. Our system performs with an accuracy of $96.489\%$, $F_1$ score of $97.334\%$, and a true positive rate (tpr) of $87.993\%$ at a remarkably low false positive rate (fpr) of $0.098\%$. This reliable and efficient system for identifying cattle can significantly advance livestock insurance and precision farming.
\end{abstract}


\end{frontmatter}

% \linenumbers

\section{Introduction}\label{intro}
The significance of cattle identification cannot be overstated. Proper identification of cattle is necessary for insurance claim verification and precision livestock farming, which includes tracking, monitoring, and maintaining production records. The conventional methods of identification, such as visual identification, neck chains, ear tags, collar tags, leg tags, and tattoos have their own drawbacks like data tampering, health problems, and high costs~\citep{kaur2022cattle}. We have implemented a muzzle-based biometric identification system for cattle to get around those limitations.

The muzzle (\textit{Planum nasolabiale}) is the fusion of the nasal entrance and upper lip in cattle. It protrudes a hard structural component, a wide and wet surface. The patterns on every cattle muzzle are distinctive~\citep{petersen1922identification, gilmore1950inherited, bello2020cattle}, just like a person's fingerprints. This scientifically proven observation has been used for cattle identification since 1922. At that time, ink impressions and hand paintings of muzzles were used for cattle identification. Digital cameras are now used to capture the muzzle pattern, and machine learning~\citep{domingos2012few} and deep learning~\citep{lecun2015deep} algorithms are then applied to the images to extract useful information. These algorithms have great implications in facial recognition which finds more precise characteristic patterns from an image~\citep{li2022individual}.

Face recognition technology uses a digital image or a video frame from a video source to automatically identify or verify a person~\citep{chellappa1995human, bhattacharyya2009biometric}. For face recognition, numerous methods and approaches have been proposed~\citep{wang2021deep}. In the early 1990s, after the historical eigenface~\citep{turk1991eigenfaces} approach was introduced, face recognition research gained popularity. Despite opening up a brand-new area of study, face recognition technology was not progressing promptly until AlexNet~\citep{10.5555/2999134.2999257} won the ImageNet~\citep{deng2009imagenet} competition in 2012. Deep learning methods, such as Convolutional Neural Network (CNN)~\citep{lecun1998gradient}, Long Short-Term Memory (LSTM)~\citep{hochreiter1997long}, Transformer~\citep{vaswani2017attention}, and others use a cascade of multiple layers of processing units for feature extraction and transformation. These internal layers learn multiple levels of representations that correspond to various degrees of abstraction~\citep{zeiler2014visualizing}. The levels make up a hierarchy of concepts that, when it comes to face recognition, exhibits strong invariance to changes in lighting, expression, pose, and other factors. In 2014, DeepFace~\citep{taigman2014deepface} achieved state-of-the-art accuracy of $97.35\%$ on the famous Labeled Faces in the Wild (LFW)~\citep{huang2008labeled} benchmark, approaching the average performance of human on facial recognition in unconstrained condition. And the very next year FaceNet~\citep{schroff2015facenet} was introduced with a record-breaking accuracy of $99.63\%$ on the LFW benchmark.

The majority of the above-mentioned technologies are frequently used in the field of human face recognition. In addition, these technologies have also shown respectable performance when they are used in the identification of cattle by its face or muzzle~\citep{kaur2022cattle}. \citet{cai2013cattle} used a facial representation model of cattle based on Local Binary Pattern (LBP)~\citep{brahnam2014local} texture features and some extended LBP descriptors. \citet{tharwat2014cattle} extracted local invariant features from muzzle print images using LBP. They also applied different machine learning algorithms such as Naive Bayes, Principal Component Analysis (PCA), K-Nearest Neighbor (KNN), and others for identifying individual cattle. \citet{mahmoud2015automatic} and \citet{sian2020cattle} used Support Vector Machine (SVM)~\citep{cortes1995support} for a muzzle-based classification system. \citet{kumar2017muzzle} also used the texture feature descriptors, such as speeded-up robust feature and LBP for the extraction of features from the muzzle point images. \citet{shojaeipour2021automated}, \citet{xu2021evaluation}, and \citet{kaur2022cattle} used various types of CNN-based architectures such as ResNet101, VGG19, DenseNet121, InceptionV3, and others with transfer learning for modeling the problem of identifying individual cattle as a classification task. \citet{mcdaid2022using} created a dataset containing 32 individual cows and applied a convolutional siamese neural network for the facial recognition of cattle.

The majority of the formerly suggested cattle identification systems employ a classification layer at the very end and trained the model using a small number of cattle images. In this paper, we have proposed, a practical method to solve the cattle identification problem by training a FaceNet architecture with a self-assembled cattle muzzle dataset to learn a meaningful feature representation. To extract the muzzle portion from the cattle image, an object detection algorithm~\citep{zou2019object} was used, followed by the FaceNet.

\section{Methodology}
\subsection{Data Collection}
\begin{figure}[h]
    \centering
    \includegraphics[width=1.0\linewidth]{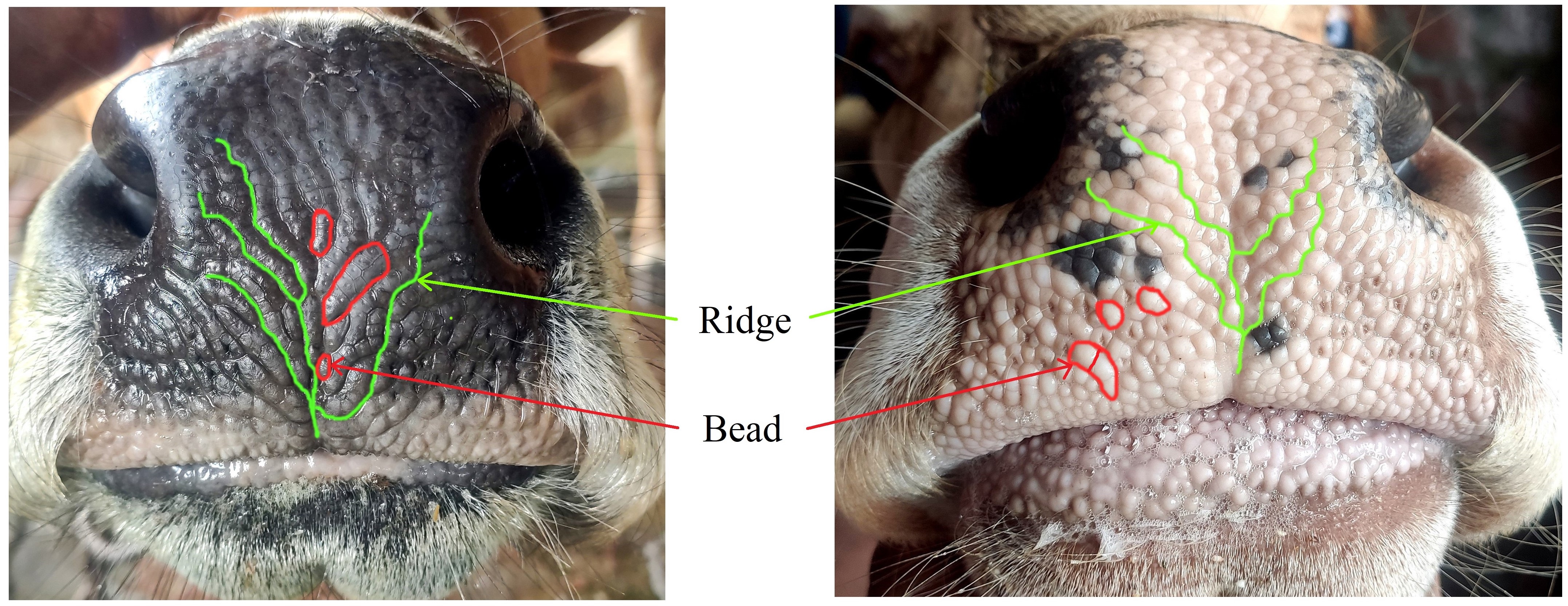}
    \caption{The configuration of the ridges and the beads.}
    \label{fig:fig_01}
\end{figure}

Data is a crucial component in machine learning. Having enough diversity in the dataset ensures that models learn properly and produce concrete results. Again, an inaccurate, biased, or incomplete dataset can lead to incorrect predictions. A high-quality dataset was required to solve the cattle identification problem. In 2021–2022, there were about 24.7 million\footnote{\href{http://dls.gov.bd/site/page/22b1143b-9323-44f8-bfd8-647087828c9b/Livestock-Economy}{``Livestock Economy at a glance 2021-2022'', published by the Department of Livestock Services (DLS) of Bangladesh.}} cattle in Bangladesh which presented a great opportunity to create a dataset of cattle muzzles from this large cattle population. Data was collected from Central Cattle Breeding and Dairy Farm\footnote{\href{http://ccbs.gov.bd}{Central Cattle Breeding and Dairy Farm is a Bangladesh government-owned cattle breeding farm and research center. It is located at Savar, Dhaka.}} and adorsho praniSheba Ltd R\&D Farm\footnote{adorsho praniSheba Ltd R\&D Farm is a model farm that is used for research and testing of innovative technologies.}. When collecting muzzle images, the data collection team took great care to ensure the image quality. The main concerns were mostly about whether the cattle muzzle ridges and beads (shown in Fig.~\ref{fig:fig_01}) were clear, as well as the clarity and focus of the image at the proper angle to allow for the retrieval of the most information possible.

\vspace{3em}

A total of 826 individual cows' image data was collected. The majority of the cattle, which ranged in age from 6 months to 8 years, were mostly of the Holstein Friesian, Sahiwal, Sindhi, and native breeds of Bangladesh. Due to the fact that the majority of the data collection was done on dairy farms, there were a lot of female cattle in the dataset, with a ratio of about 4:1 to male cattle. During the collection process, approximately 40,000 raw images were collected in total. Two experts manually reviewed those images, and a handful number of them were eliminated from the dataset due to unsatisfactory quality. After the removal of those images, a total of 32,374 images remained in the dataset.

All of the pictures were taken with various smartphone cameras. Two skilled data annotators work together to label the complete set of data. To determine the inter-rater reliability of the cleanness and annotation, a skilled dermatologist examined a portion of the annotated images. Cohen's kappa statistic~\citep{cohen1960coefficient} was used to measure the inter-rater agreement and revealed strong agreement ($\kappa=0.72$) between the two annotators.

\begin{figure}[h]
    \centering
    \includegraphics[width=0.99\linewidth]{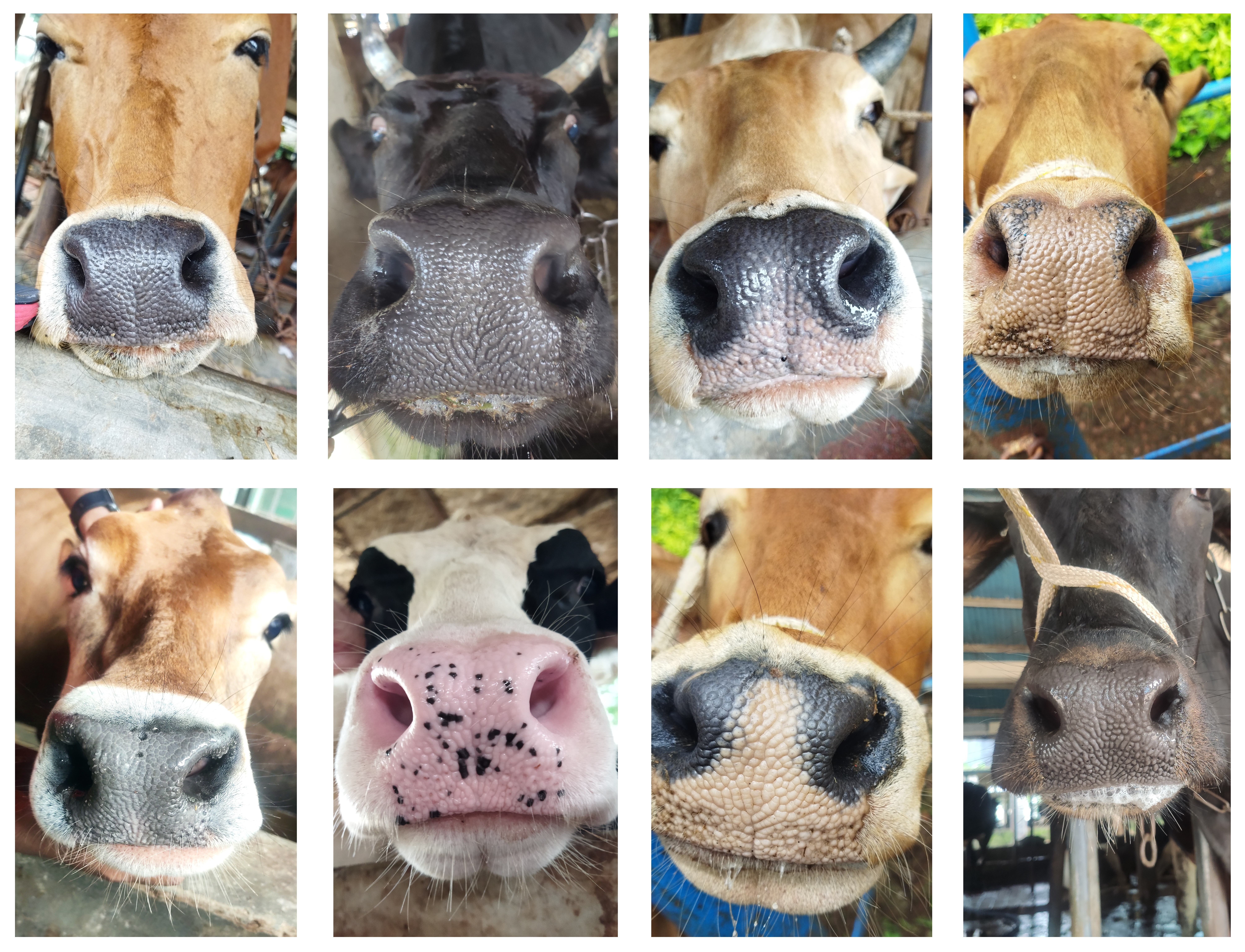}
    \caption{Sample images from our dataset.}
    \label{fig:fig_02}
\end{figure}

\vspace{3em}

\subsection{Cattle Muzzle Detector}
Recognition algorithms rely heavily on the region of interest they operate on. Thus, a good object detection module is crucial for successful recognition in any identification system pipeline. Object detection algorithms have generally been used to accurately locate regions of interest~\citep{girshick2015region}.
 
The YOLO algorithm was initially proposed by \citet{redmon2016you}. It is one of the few algorithms that has shown fast response and precision in finding multiple objects in a single image. YOLO divides an image into a number of grids and predicts bounding boxes and confidence scores of a particular object or region of interest simultaneously in all the grids. A bounding box is represented by four parameters: the x and y coordinates of the box's center, as well as its width and height. These parameters define the position and size of the image's bounding box. The confidence score, also known as the objectness score, expresses the YOLO algorithm's belief that the bounding box contains an object of interest. The confidence score is based on the likelihood of an object being present in the bounding box, the prediction's accuracy, and the overlap with other predicted bounding boxes.

\begin{figure}[h]
    \centering
    \includegraphics[width=1.0\linewidth]{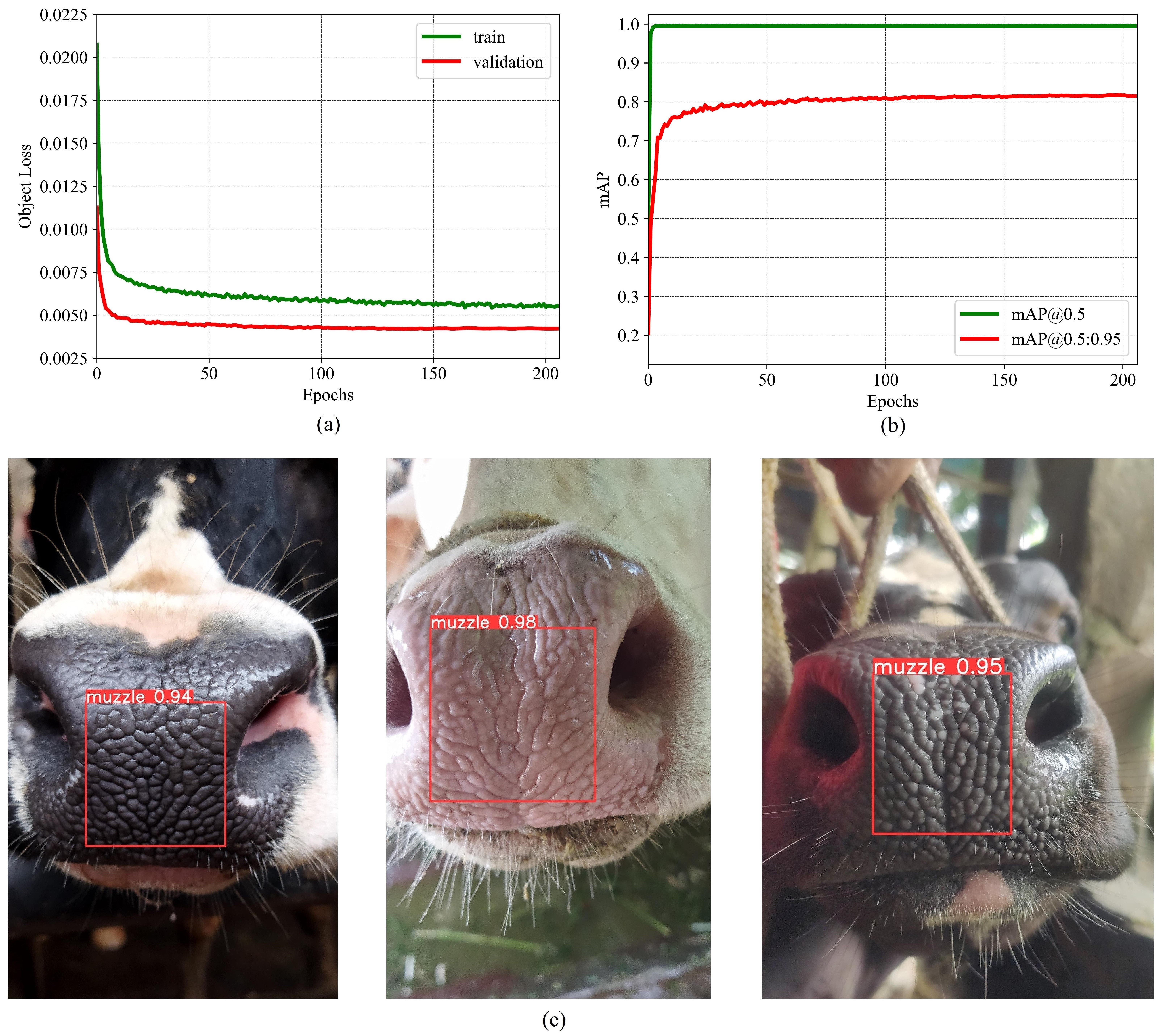}
    \caption{(a) Object loss of YOLOv5 model, (b) Mean average precision of YOLOv5 model, (c) Sample of muzzle detection}
    \label{fig:fig_03}
\end{figure}

Our system seeks to recognize a cow based on the pattern of its muzzle. As a result, the muzzle must be extracted from a full-face image of a cow because it is the area we are interested in. To serve this purpose, we have used YOLO to crop the muzzle for this reason in order to propagate it to the FaceNet section. To train YOLO, around 5,000 images of cattle were annotated. Two annotators manually annotated the muzzle portion from each image. These images were split into a 4:1 ratio to build up the train and test sets. YOLOv5 was trained for 207 epochs. Following training, it achieved mAP@0.5 of 0.99 and mAP@0.5:0.95 of 0.82 with significantly low training and validation loss. It had a classification accuracy of $98\%$ during the testing period. Fig.~\ref{fig:fig_03} shows the performance of the YOLOv5 model on muzzle detection.

\subsection{Data Preprocessing}
Every image went through two preprocessing steps: Sharpening, and Adaptive Histogram Equalization. The followings describe the two methods.

Sharpening: Sharpening is a process of enhancing the edge of an image. An edge refers to a significant change in image intensity or brightness. A sharpened image has a higher resolution and more discernible details~\citep{archana2016review}. These enhanced edges provide useful information that plays a significant role in extracting features from an image by a deep neural network. \citet{gonzales2018digital} stated the sharpening operation with the following equation: 
\[
    g(x,y) = f(x,y) + k*g_{mask}(x,y)
\]

Where, $f(x,y)$ is the original image, $g(x,y)$ is the sharpened image, $g_{mask}(x,y)$ is obtained by subtracting a blurred version of an image from its original version, and $k$ is a weight factor that determines the degree of sharpness. To find this mask, we have used a weighted median filter, 

\[
\begin{bmatrix}
0 & -1 & 0\\
-1 & 5 & -1\\
0 & -1 & 0
\end{bmatrix}
\]

Convolving the original image with this filter obtains a blurry version of an image, which is furthermore used to find the unsharp mask.

Adaptive Histogram Equalization: To increase image contrast, a well-known technique called histogram equalization (HE) is used. With this method, pixels are essentially redistributed so that their cumulative density function exhibits a linear trend after computing the histogram of pixel intensity for an image.

HE increases global contrast, which may have a negative impact on how well local contrast is improved~\citep{xie2019image}. Histogram equalization has been improved with Adaptive Histogram Equalization (AHE). The AHE algorithm divides an image into multiple grids and generates a unique histogram for each grid individually. As a result, AHE enhances the local contrasts of an image but also overamplifies the noises of that image. 

\begin{figure}[h]
    \centering
    \includegraphics[width=0.6\linewidth]{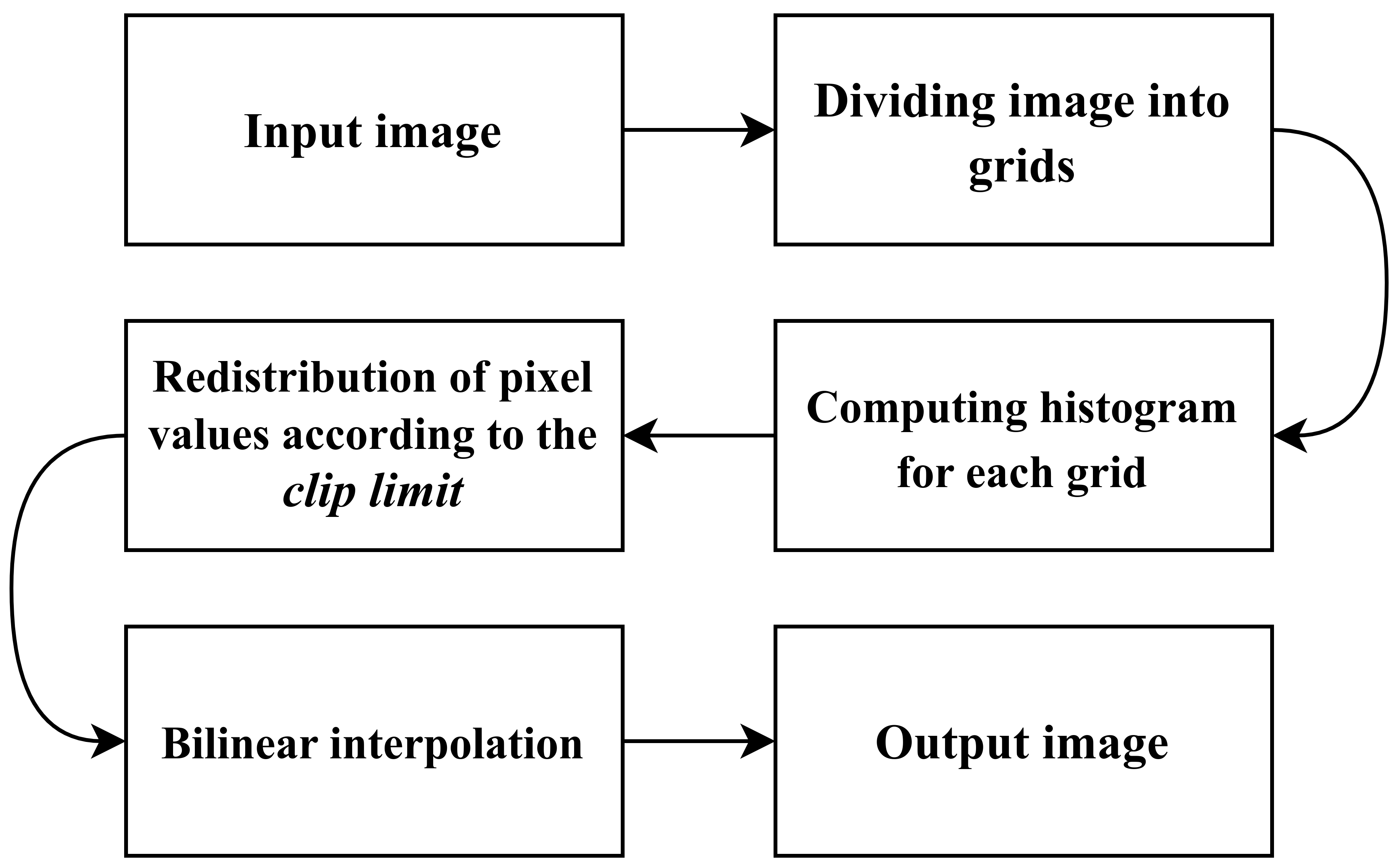}
    \caption{Image enhancement procedure by CLAHE.}
    \label{fig:fig_04}
\end{figure}

Contrast Limited Adaptive Histogram Equalization (CLAHE) solves the noise amplification issue by placing a predefined limit, stated as \textit{clip limit}, on the amplification of contrast. As a variation of AHE, CLAHE divides an image into grids and finds histograms for each segment. The pixel values that go over that contrast threshold are redistributed to the other pixel-intensity bins for each histogram. After that, the segments of the image are tied up through bilinear interpolation. Fig.~\ref{fig:fig_04} illustrates this procedure.

\subsection{Feature Extraction}
\subsubsection{Architecture}
\begin{figure}[h]
    \centering
    \includegraphics[width=1.0\linewidth]{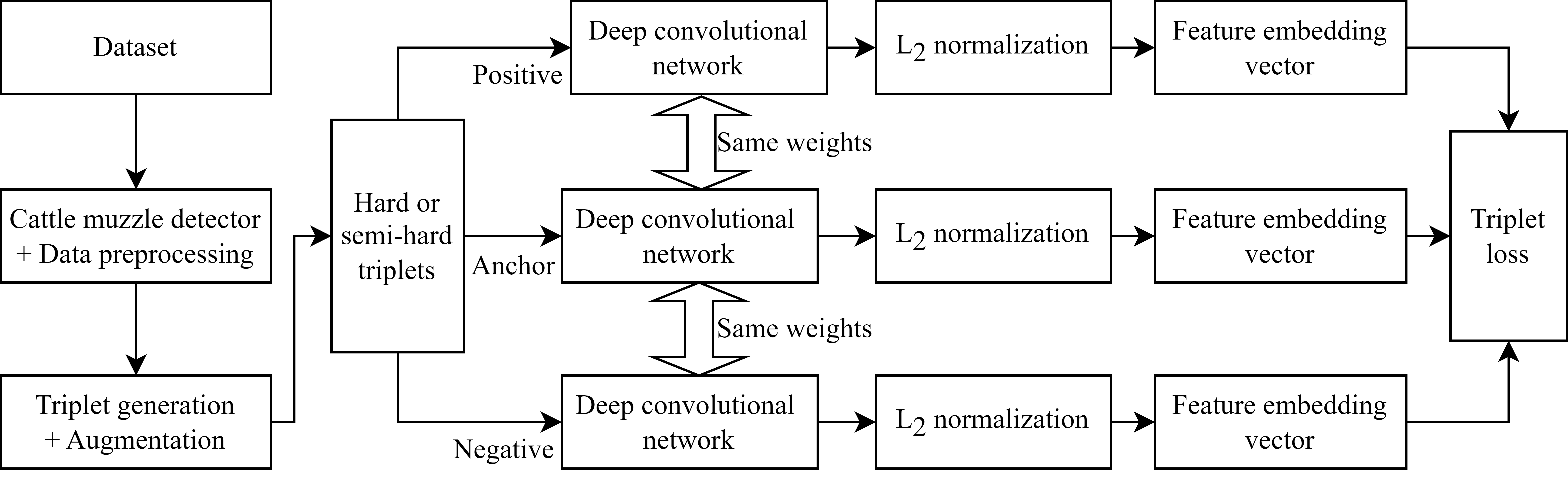}
    \caption{Overview of feature extraction architecture using FaceNet.}
    \label{fig:fig_05}
\end{figure}

The majority of face recognition and verification systems based on deep learning operate in a classification-like fashion~\citep{sun2015deeply, taigman2014deepface}. With more classes, this becomes less accurate and computationally expensive while also becoming less effective. \citet{schroff2015facenet} proposed a unified face verification, recognition, and clustering system named FaceNet that outperformed traditional approaches by setting a new benchmark. Our system, having FaceNet architecture as the core part, created a euclidean space where a point represents each individual muzzle. Squared $L_2$ distance between any two points represents the dissimilarity between the two muzzles that were used to make these points of embedding. We aimed to minimize these squared distances among the muzzles of the same cattle and maximize distances among different cattle; for achieving this, ``Triplet Loss'' was used. A deep convolutional neural network is needed to be used in the FaceNet architecture for extracting features from an image by passing it through the convolutional layers. We tried out several well-explored deep networks like ResNet50~\citep{he2016deep}, InceptionV3~\citep{szegedy2016rethinking}, VGG16~\citep{simonyan2014very}, MobileNetV2~\citep{sandler2018mobilenetv2}, DenseNet169~\citep{huang2017densely} and among them ResNet50 was selected as it performed better. Fig.~\ref{fig:fig_05} illustrates an overview of the architecture of the entire muzzle-based cattle identification system.

\subsubsection{Triplet Generation}
FaceNet embeds an image $x$ to a feature vector $f(x)$. Any muzzle image $x_a^i$ of a specific cow can be considered as the anchor. \(T^i = \{ f(x_a^i), f(x_p^i), f(x_n^i) \} \) is considered as $i^{th}$ triplet where $x_p^i$ (positive) is another muzzle image of that cow, and $x_n^i$ (negative) is a muzzle image of a different cow. If the distance between the positive image pair is $d_p^i$, and the distance between the negative image pair is $d_n^i$, then the triplet loss $L$ would be defined as,

\[
L = \sum_{i=0}^{N} \Big[ d_p^i - d_n^i + \alpha \Big]_+
\]

Here, \( d_p^i = \vert\vert f(x_a^i) - f(x_p^i) \vert\vert_2 ^2 \), \( d_n^i = \vert\vert f(x_a^i) - f(x_n^i) \vert\vert_2 ^2 \), and $\alpha$ is a hyperparameter that draws the marginal line between the positive image pair and the negative image pair.

$T^i$ can be defined as an easy triplet, semi-hard triplet, or hard triplet for holding the following conditions \( (d_n^i - d_p^i) \ge \alpha \),  \( 0  <  (d_n^i - d_p^i) < \alpha \), or   \( (d_n^i - d_p^i) \le  0 \) respectively.

\begin{figure}[h]
    \centering
    \includegraphics[width=0.6\linewidth]{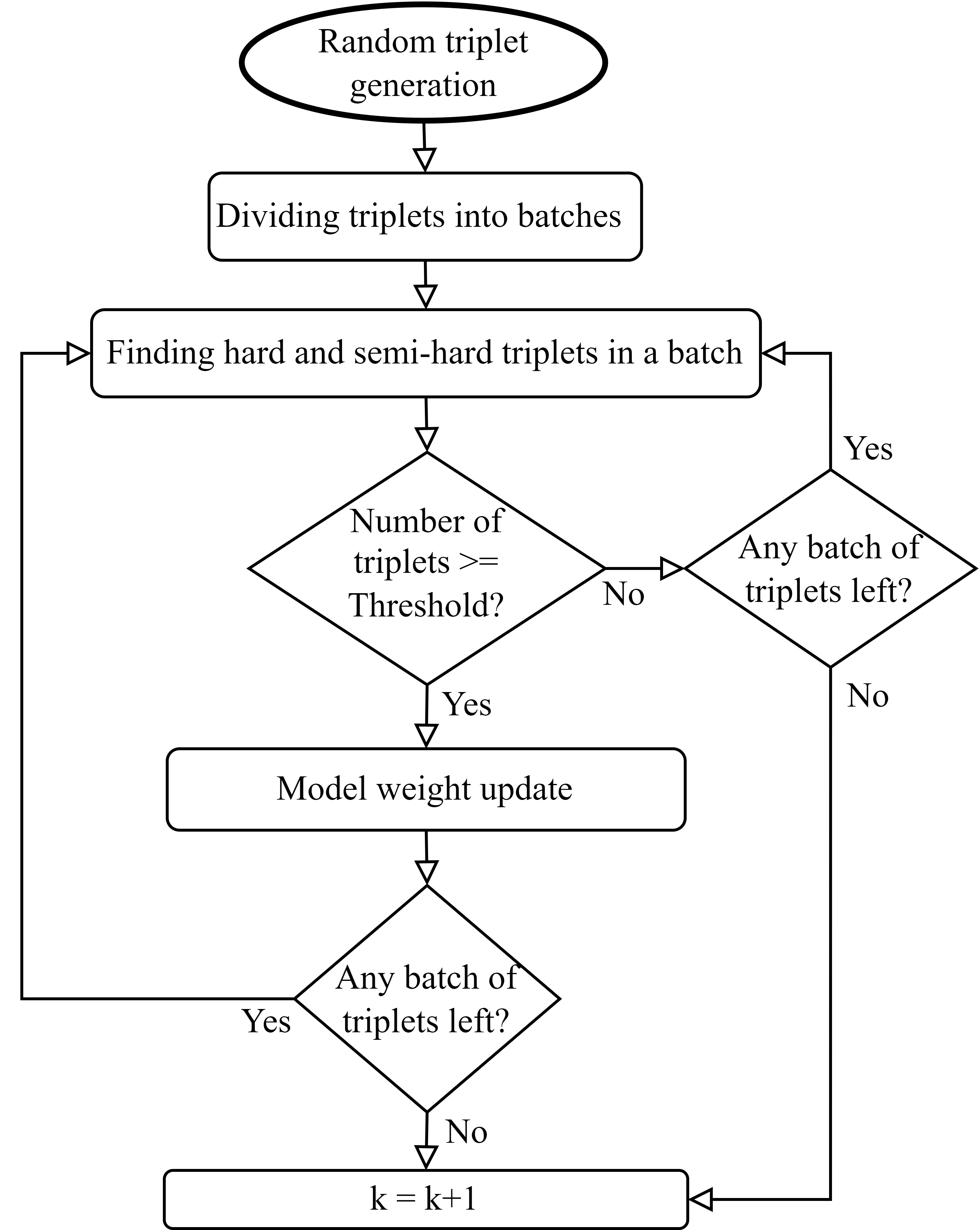}
    \caption{Triplet generation procedure for $k^{th}$ epoch.}
    \label{fig:fig_06}
\end{figure}

\begin{figure}[h]
    \centering
    \includegraphics[width=0.6\linewidth]{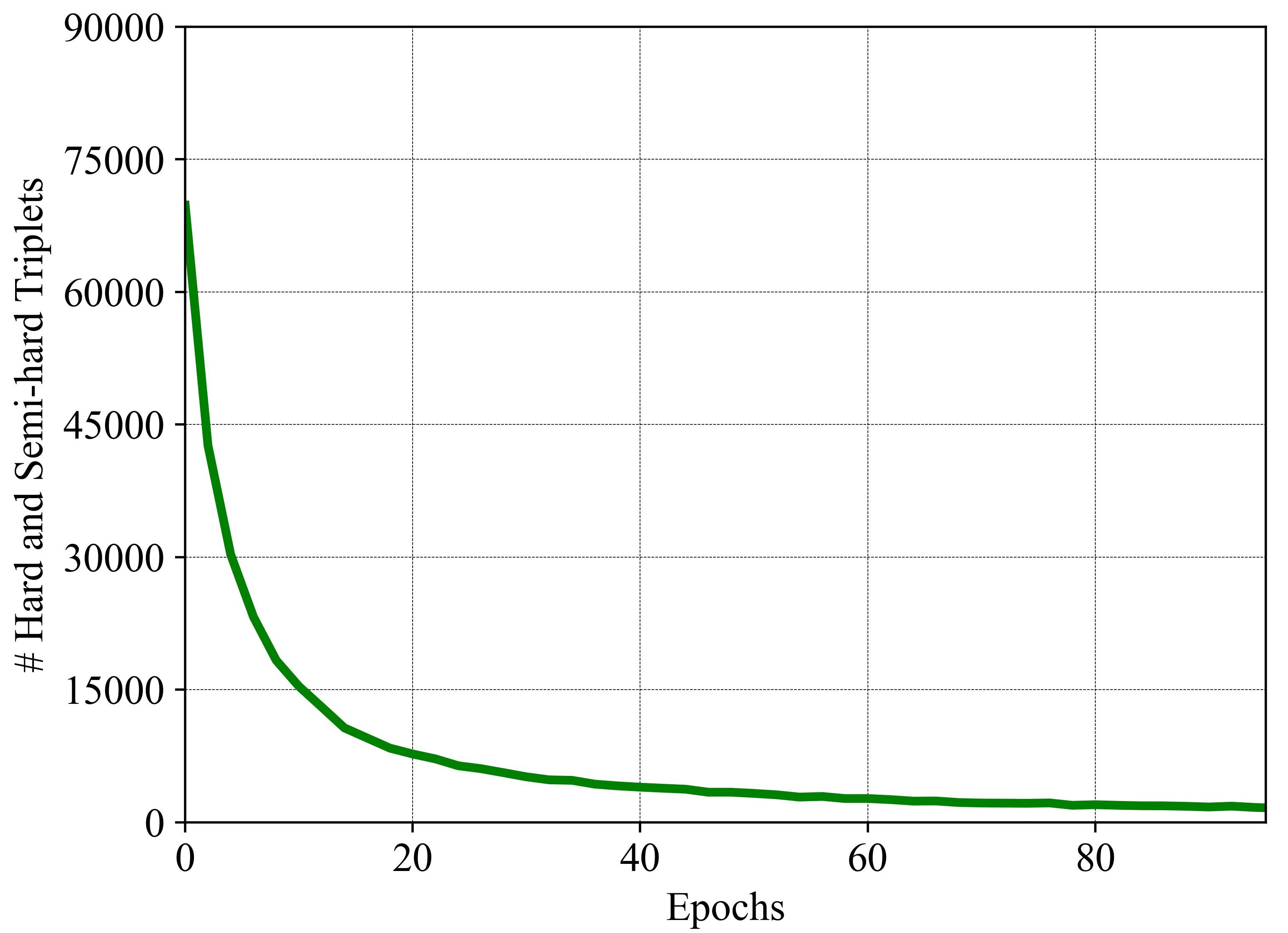}
    \caption{Number of hard and semi-hard triplets found in each epoch.}
    \label{fig:fig_07}
\end{figure}

Fig.~\ref{fig:fig_06} describes our triplet generation procedure. Since only hard triplets may result in local minima and easy triplets do not contribute to the training process, we have chosen a semi-hard and hard triplet generation process for triplet selection in each iteration. For each epoch, model weights from the previous update were used to find semi-hard and hard triplets. We set a minimum threshold for the number of triplets; model weights were not updated until the number of triplets reached that threshold. This ensured maximizing the advantages of mini-batch gradient descent by minimizing the oscillation of loss function during training~\citep{lecun2002efficient}. The number of semi-hard and hard triplets was found to be reduced as the model got to be more mature, which can be seen in Fig.~\ref{fig:fig_07}.

\subsubsection{Image Augmentation}
In the training phase, generated triplets were augmented using several standard image augmentation techniques such as rotation, zooming, cropping, shearing, translation, and random horizontal and vertical flipping. Most of the machine learning frameworks have support for these augmentations. In this experiment, the PyTorch\footnote{\href{https://pytorch.org/}{PyTorch is a machine learning framework based on the Torch library, used for applications such as computer vision and natural language processing, originally developed by Meta AI and now part of the Linux Foundation umbrella.}} framework was used. Other than these conventional techniques, two more techniques were used which are the addition of blackout, and salt and pepper noise.

Blackout: Randomly selected rectangular or square portions of the image were converted to a modified area i.e. pixels of that area were converted to random pixel values. A random rectangular or square area of image $F_{M \times N}$, whose coordinates constructed the set $C_{blackout}$, was selected. Augmented image, 
\[
F_{aug} = (F \odot B) + C
\]

Here, $C_{blackout}$ is a set constructed by the coordinates of randomly selected rectangular or square portions of the image to modify that area to a random pixel value.

\vspace{-2.5em} 

\[
\begin{aligned}
B_{(M \times N)} &= [b_{(i,j)}]\quad where\ b_{(i,j)} = 0\quad if\ (i, j) \in C_{blackout},\ or\ 1\  otherwise \\
C_{(M \times N)} &= r \times [c_{(i,j)}]\quad where\ c_{(i,j)} = 1\quad if\ (i, j) \in C_{blackout},\ or\ 0\  otherwise, \\ 
and\quad r &\sim U(0, 255)
\end{aligned}
\]

Salt and Pepper Noise: Salt and Pepper noise occurs in the form of some black or white pixels on the digital image. After adding noise to image $F_{M \times N}$, augmented image, 

\[
F_{aug} = ((F \odot S) + R) \odot P
\]

Here, $C_{salt}$ and $C_{pepper}$ are two sets constructed by randomly selected coordinates of the image where salt and pepper are to be added respectively.

\vspace{-2.5em} 

\[
\begin{aligned}
S_{(M \times N)} &= [s_{(i,j)}]\quad where\ s_{(i,j)} = 0\quad if\ (i, j) \in C_{salt},\ or\ 1\  otherwise \\
R_{(M \times N)} &= [r_{(i,j)}]\quad where\ r_{(i,j)} = 255\quad if\ (i, j) \in C_{salt},\ or\ 0\  otherwise \\
P_{(M \times N)} &= [p_{(i,j)}] \quad where\ p_{(i,j)} = 0\quad if\ (i, j) \in C_{pepper},\ or\ 1\  otherwise        
\end{aligned}
\]

Some aggressive transformations were applied to the training data to produce new, slightly different training triplets while leaving the validation data unchanged which is an accurate representative sample of the real world. The model performed better on the validation data than the training data, as shown in Fig.~\ref{fig:fig_08} because the validation data was not augmented.

\section{Results}

\subsection{Experimental Setup}
We trained our model on a local machine equipped with 11th Gen Intel(R) Core\textsuperscript{TM} i5-11400 processor with a clock speed of 2.60 GHz and 32 GB RAM. We used NVIDIA GeForce RTX 3060 graphics card with 12 GB GDDR6 memory. The dataset was split into the train, validation, and test set (Table~\ref{tab:table_01}). No cattle images from the trainset were present in either the validation or test sets, which also holds true for both validation and test sets. We selected ResNet50, InceptionV3, VGG16, MobileNetV2, and DenseNet169 as base models for the FaceNet system for 5 different experiments. The final selected model, ResNet50 took approximately 400 hours to be trained for 96 epochs. Our models were trained from scratch without any transfer learning and the model weights were stored for each epoch.

\begin{table}[h]
   \centering
   \renewcommand{\arraystretch}{1.0}
   \caption{Data partitioning}
   \label{tab:table_01}
	\begin{tabular}{|P{0.95in}|P{0.7in}|P{0.7in}|P{0.85in}|P{1.25in}|}
        \hline
		\textbf{Dataset subsection} & \textbf{\#cattle} & \textbf{\#total image} & \textbf{Coefficient of variation} & \textbf{\#randomly generated triplets}\\
        \hline \hline
		 Train set & 620 & 23,954 & $18.37\%$ & 1,00,000 \\
         Validation set & 164 & 6,410 & $18.52\%$ & 25,000 \\
         Test set & 42 & 2,010 & $17.93\%$ & NA \\
        \hline
	\end{tabular}
\end{table}

\subsection{Performance Analysis}
\begin{table}[ht]
   \centering
   \renewcommand{\arraystretch}{1.0}
   \caption{Performance comparison}
   \label{tab:table_02}
	\begin{tabular}{|P{1.0in}|P{0.9in}|P{0.8in}|P{1.0in}|}
        \hline
		\textbf{Model} & \textbf{\#parameters} & \textbf{Accuracy} & \textbf{VAL\tablefootnote{Reported is the mean Validation Rate (VAL) with standard deviation at $10^{-3}$ False Accept Rate (FAR). VAL and FAR are defined in section 4 of \citet{schroff2015facenet}.}}\\
        \hline \hline
		ResNet50 & 36.35M & $96.49\%$ & $83.37\% \pm 0.54$ \\
        InceptionV3 & 28.36M & $96.25\%$ & $83.19\% \pm 0.51$ \\
        VGG16 & 17.93M & $92.72\%$ & $81.39\% \pm 0.63$ \\
        MobileNetV2 & 10.28M & $90.45\%$ & $80.55\% \pm 0.71$ \\
        DenseNet169 & 23.08M & $92.18\%$ & $82.26\% \pm 0.57$ \\
        \hline
	\end{tabular}
\end{table}

Table~\ref{tab:table_02} shows the best performance of each model. Using ResNet50, the system achieved a mean Validation Rate (VAL) of $83.37\%$ at a False Accept Rate (FAR) of $10^{-3}$ for the validation set. Our experiment also indicates that model performance slightly changes depending on the dimension of the feature vector that the FaceNet extracts out of a muzzle image. Table~\ref{tab:table_03} shows that the performance of ResNet50 was maximized while the feature vector had a length of 128.

\begin{table}[h]
   \centering
   \renewcommand{\arraystretch}{1.0}
   \caption{Feature dimensionality}
   \label{tab:table_03}
	\begin{tabular}{|P{1.0in}|P{1.0in}|}
        \hline 
        \textbf{Length of feature vector} & \textbf{VAL}\\
        \hline \hline
        64 & $82.64\% \pm 0.61$ \\
        128 & $83.37\% \pm 0.54$ \\
        256 & $83.19\% \pm 0.53$ \\
        \hline
	\end{tabular}
\end{table}

\begin{figure}[h]
    \centering
    \includegraphics[width=1.0\linewidth]{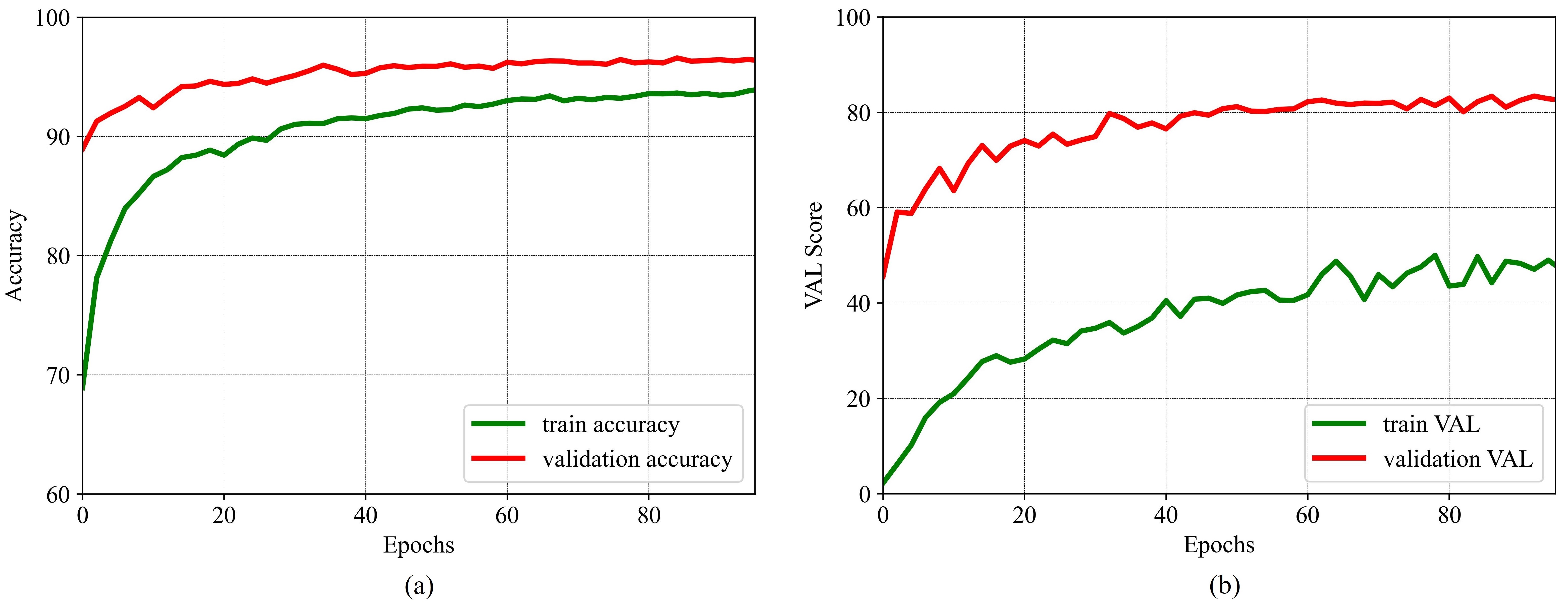}
    \caption{(a) Accuracy of the model, (b) VAL score of the model.}
    \label{fig:fig_08}
\end{figure}

Having a backbone of ResNet50 in the FaceNet architecture, the model was trained with a learning rate of 0.003 which decayed $20\%$ per 8 epochs, for a total of 96 epochs. Adam optimizer was used to optimize the triplet loss function that had a 0.5 margin. Fig.~\ref{fig:fig_08} shows the performance of the model on both the training and validation datasets.

\begin{figure}[h]
    \centering
    \includegraphics[width=0.65\linewidth]{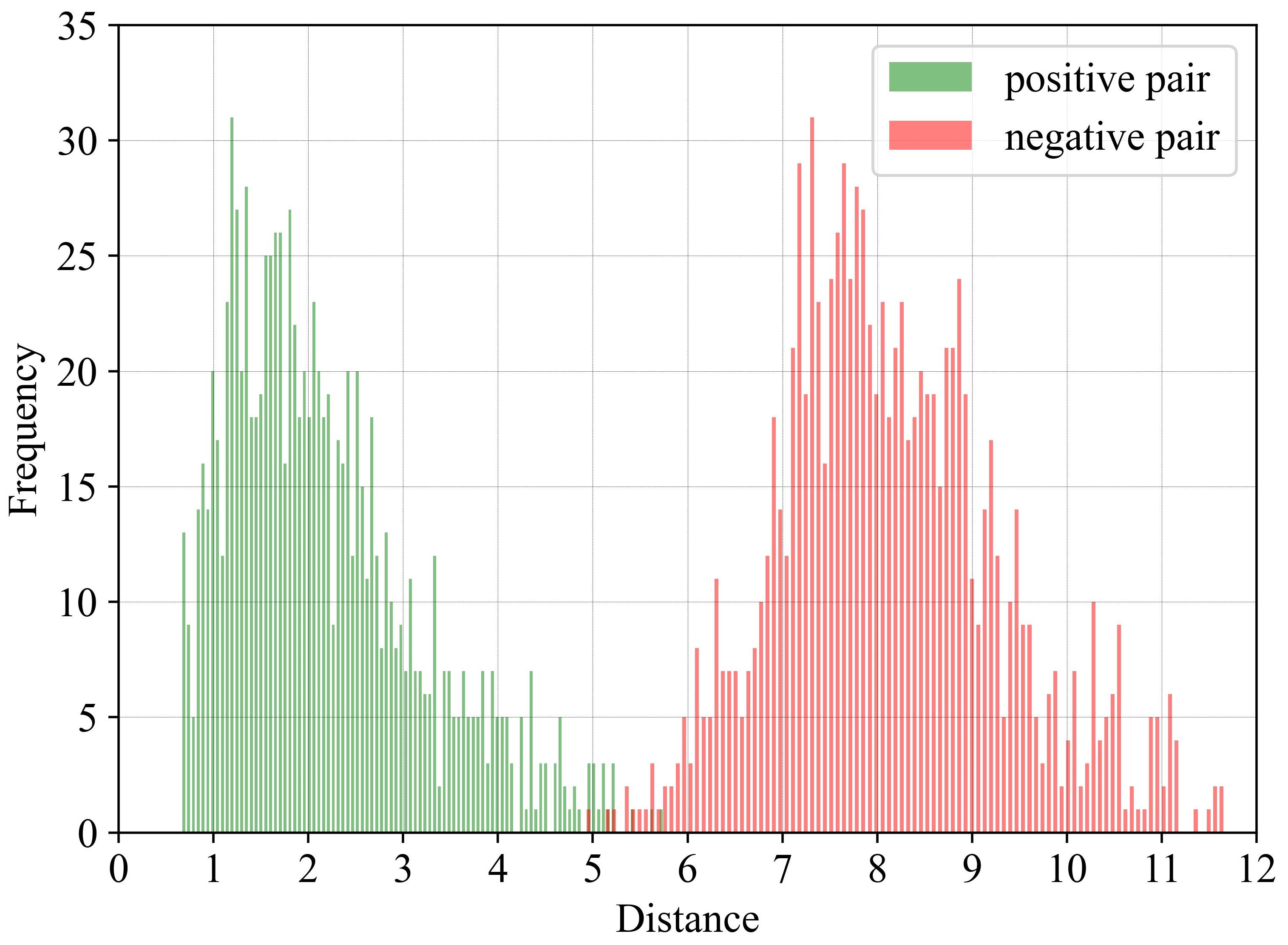}
    \caption{Distribution of positive and negative distances.}
    \label{fig:fig_09}
\end{figure}

For the test set, we had an accuracy of $96.48\%$, an $F_1$ score of $97.33\%$, and a true positive rate of $87.99\%$ at a false positive rate of $0.098\%$. To visualize the system’s performance, randomly 995 positive pairs and 995 negative pairs were selected from the test set, and the distances of each image pair were calculated with the final model. In Fig.~\ref{fig:fig_09}, it is clearly understandable that the distribution of positive and negative distances is easily distinguishable.

\begin{figure}[h]
    \centering
    \includegraphics[width=0.8\linewidth]{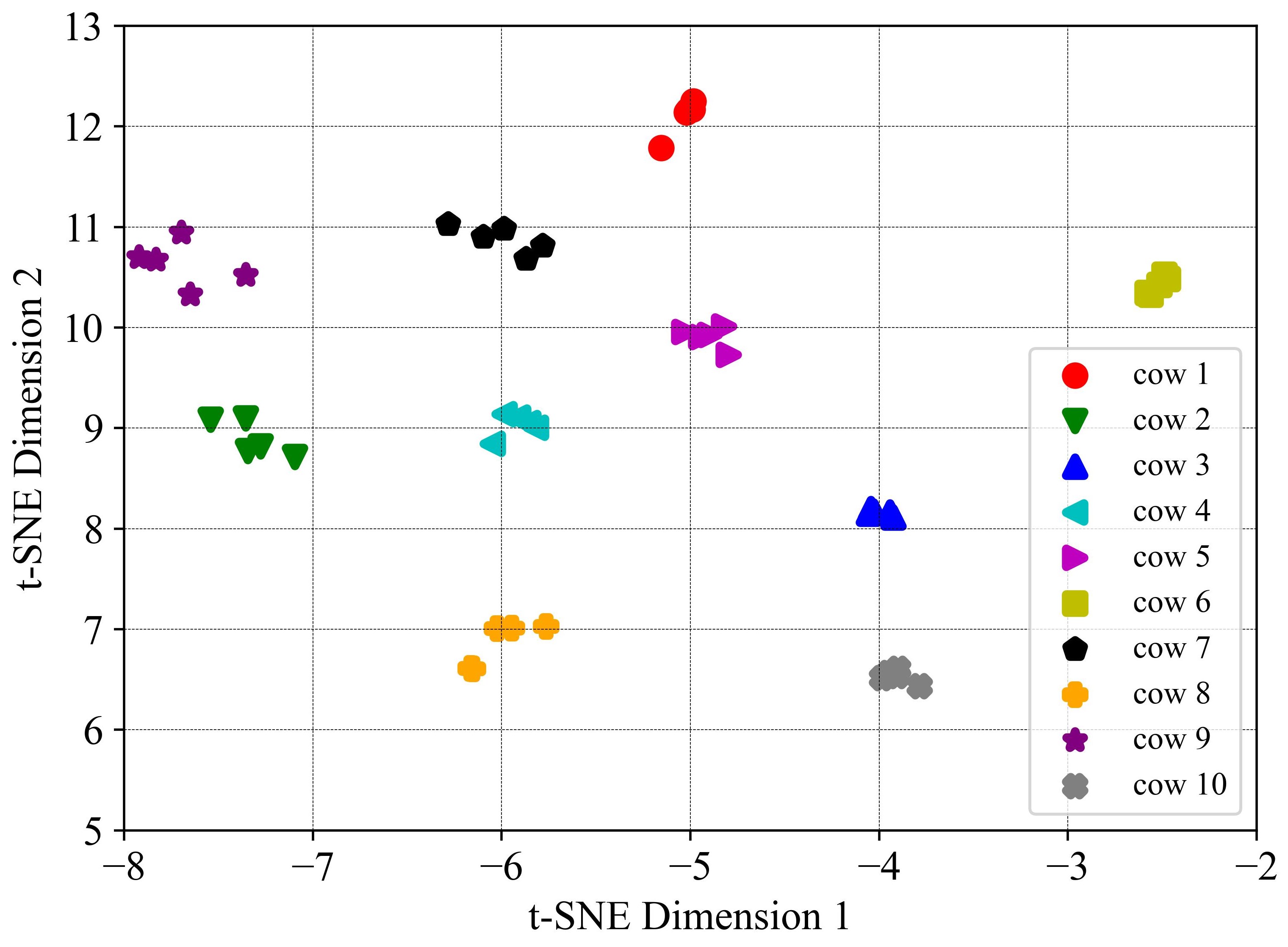}
    \caption{t-SNE visualization of cattle muzzles.}
    \label{fig:fig_10}
\end{figure}

With a random selection process, we selected 10 cattle from the test set. 5 images from each of them were picked randomly and were embedded by the model. We applied t-SNE (t-distributed stochastic neighbor embedding)~\citep{van2008visualizing} algorithm to those embeddings to model the 128-dimensional feature vectors as a set of probability distributions in a two-dimensional space. Fig.~\ref{fig:fig_10} shows a separation between distinct clusters of cattle, with a high degree of similarity within each cluster and a low degree of similarity between two different clusters.

\vspace{3em}

\section{Application}
The rural community in Bangladesh considers cattle as one of their most valuable possessions. For small, medium, and marginal farmers, rearing cattle is a substantial source of revenue. According to the Bangladesh Bureau of Statistics\footnote{\href{http://bbs.gov.bd/site/page/dc2bc6ce-7080-48b3-9a04-73cec782d0df/Gross-Domestic-Product-(GDP}{``Gross Domestic Product (GDP) of Bangladesh 2021-22'', published by BBS.}}, agriculture contributes $11.66\%$ of the GDP of Bangladesh in 2021-22. However, agriculture is particularly susceptible to natural calamities. Bangladesh is frequently listed as one of the most disaster-prone countries on the planet, with its agriculture especially vulnerable to floods, cyclones, and drought. For instance, 71.373 billion BDT was lost in the livestock sector in 2021 from disaster-related events\footnote{\href{http://bbs.gov.bd/site/page/29855dc1-f2b4-4dc0-9073-f692361112da/Statistical-Yearbook}{``Statistical Yearbook Bangladesh 2021'', published by Bangladesh Bureau of Statistics (BBS).}}. Livestock insurance is crucial for comprehensive coverage against livestock loss since farmers' livelihoods are so highly dependent on them. Several Non-Governmental Organizations (NGOs) and Microfinance Institutions (MFIs) are offering livestock-credit insurance. Nevertheless, in the absence of reinsurance agreements, these programs are not effective enough to deal with the economic loss of livestock farmers. According to a survey, no respondents had access to crop insurance, and only $1.4\%$ of 3,490 respondents had access to livestock insurance of any sort\footnote{\href{https://openknowledge.worldbank.org/handle/10986/31046}{``Bangladesh: Agriculture Insurance Situation Analysis'', published by World Bank Group.}}. Fewer than $2\%$ to $3\%$ of owners of livestock are estimated to have microcredit insurance for their cattle\footnote{\href{https://openknowledge.worldbank.org/handle/10986/31044}{``Bangladesh: Policy Options for Agriculture Insurance'', published by World Bank Group.}}. 

Any system that offers services for identifying individual cattle can incorporate our suggested method. We have developed a demo application with two key attributes— cattle enrollment and verification—to illustrate this. A cattle is enrolled into the system by having its feature embeddings discovered using the suggested model. As our approach primarily constitutes a one-shot learning~\citep{koch2015siamese} process, enrollment into the system necessitates the use of only a single image. On the other hand, the Verification procedure can successfully confirm a specific subject's identity.

For both enrollment and verification of a cow’s identity, a muzzle identification module in the application checks if an image received in the request body contains exactly one muzzle. If so, YOLO crops the muzzle portion from that cattle image and checks whether the dimension of the cropped image is acceptable for extracting meaningful features. After that, this cropped image is preprocessed with sharpening filters and CLAHE. FaceNet receives this preprocessed image and extracts unique embedding for that cow. During enrollment, a muzzle image is embedded in its 128-dimensional feature vector, and this vector is stored in the database alongside other necessary information like the corresponding cow’s unique identification number, breed type, gender type, date of birth, disease history, vaccine history etc. 

\begin{figure}[h]
    \centering
    \includegraphics[width=1.0\linewidth]{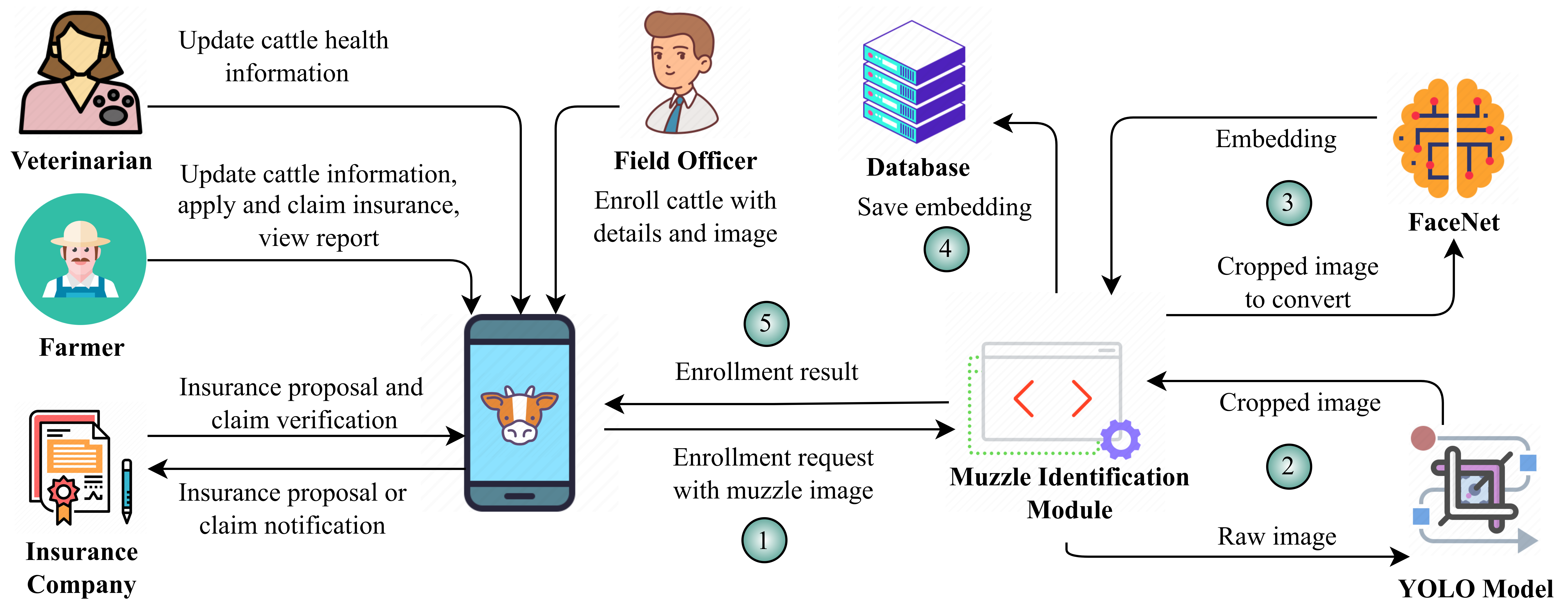}
    \caption{High-level diagram of the cattle enrollment process.}
    \label{fig:fig_11}
\end{figure}

\begin{figure}[h]
    \centering
    \includegraphics[width=1.0\linewidth]{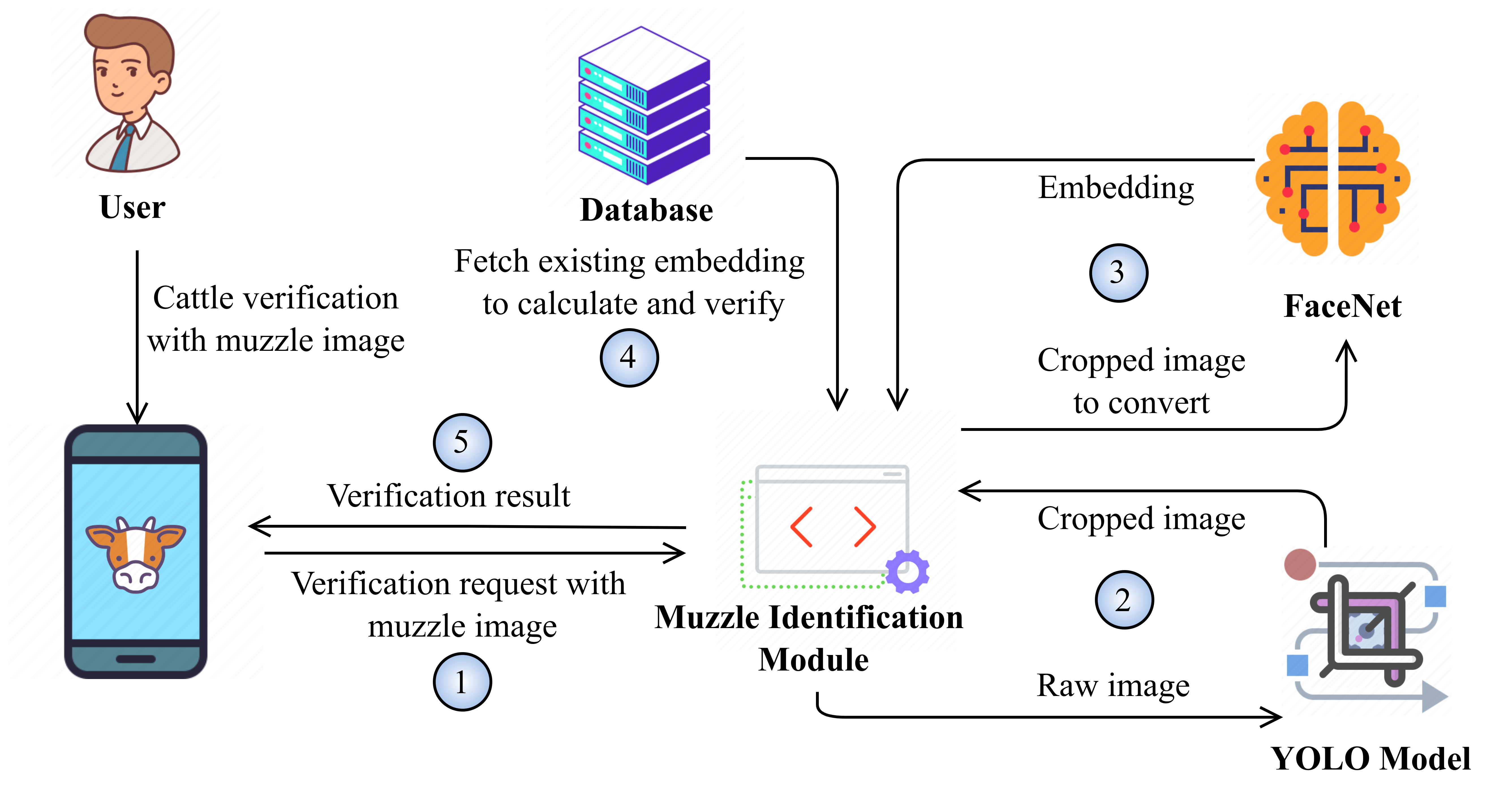}
    \caption{High-level diagram of the cattle verification process.}
    \label{fig:fig_12}
\end{figure}

Similarly, when a client requests to verify the identity of a cow, this demo application compares that cow's embedding with the existing enrolled cattle embeddings in the database. After computing distances from the embeddings, it verifies the identification of the cow based on a predetermined threshold and delivers the report to the client. Fig.~\ref{fig:fig_11} and Fig.~\ref{fig:fig_12} illustrate the real-life application of our muzzle-based cattle identification system.

\vspace{1em}
\section{Conclusion}
The proposed muzzle-based cattle identification system might revolutionize the livestock industry. This technology can demonstrate significant potential by facilitating the flow of information and access to extensive services for farmers. By accurately identifying individual animals, farmers can track their performance and provide targeted care and services. This system can also be used to facilitate access to markets, information, and entrepreneurial opportunities. For example, this tamper-proof technology can provide affordable and accessible insurance for farmers. This can help mitigate risks associated with livestock farming and provide farmers with greater financial security.

There are exciting opportunities for further enhancing this study. Our data collection primarily focused on dairy farms, resulting in a dataset of cattle all under a specific age (8 years in our case). To improve data distribution, introducing more diversity in the ages of cattle could be considered. Additionally, modern dairy farms often employ Artificial Insemination, leading to a dataset skewed towards female cattle. Future work may explore the inclusion of more male cattle to achieve a balanced representation. Although our model exhibited impressive performance in identifying cattle from clean images, it is partially affected by factors such as lighting conditions during image capture, muzzle cleanliness, and image quality. Future researchers can explore innovative techniques to address these challenges and further enhance our model's capabilities. With further research and development, this technology can also be used to identify buffaloes as their muzzles are also unique. Continued research in this area can help to unlock the full potential of this system.

While there are still challenges that need to be addressed such as the little availability of internet connections and reluctance to use modern technologies in underdeveloped areas, the potential benefits of a muzzle-based cattle identification for livestock insurance are undeniable. The Government of Bangladesh and insurance companies are getting more interested in offering livestock insurance since this technology makes the cattle identification process more convenient than ever. This implementation of the digital inclusion initiative is in the expansion of innovative digital services, which provide solutions to poor families is also in line with the Sustainable Development Goals (SDGs) 2030\footnote{\href{https://sdgs.un.org/2030agenda}{``Transforming our world: the 2030 Agenda for Sustainable Development'', published by United Nations.}}.

Overall, the muzzle-based cattle identification system is a powerful tool for improving the productivity and profitability of livestock farming. By providing farmers with accurate and timely information can help them make more informed decisions and improve the overall management practices in Bangladesh and beyond.

\section*{Acknowledgments}
The authors would like to express their sincere gratitude to adorsho praniSheba Ltd\footnote{\href{https://pranisheba.com.bd}{adorsho praniSheba Ltd is an AgriTech and InsurTech outfit based in Dhaka, Bangladesh. Its focus is to find and implement technology-driven sustainable solutions that increase the availability of financial access to agricultural entrepreneurs.}} for funding the cattle identification project and to the Department of Livestock Services (DLS) of Bangladesh for giving official permission to collect invaluable data from Bangladesh Central Cattle Breeding and Dairy Farm.

\newpage
%\bibliographystyle{elsarticle-harv}  %author-year citation style
%\bibliographystyle{elsarticle-num-names}  %number and names citation style
%\bibliographystyle{elsarticle-num} %number citation style
%\bibliography{bibfile}

\end{document}